\documentclass[runningheads]{llncs}
\usepackage{graphicx}
\usepackage{listings} % For pseudocode
\usepackage{lipsum}   % For generating dummy text, can be removed in your actual document
\usepackage{amsmath}
\usepackage{amssymb}
\usepackage{booktabs} % For prettier tables
\usepackage{siunitx} % For aligning numbers by decimal point
\usepackage{gensymb}
\usepackage{siunitx}
\usepackage{url}
\usepackage[linesnumbered,ruled,vlined]{algorithm2e}

\begin{document}
\title{Compute-Efficient Medical Image Classification with Softmax-Free Transformers and Sequence Normalization}
%
%\titlerunning{Abbreviated paper title}
% If the paper title is too long for the running head, you can set
% an abbreviated paper title here
%
\author{
Firas Khader\inst{1} \and
Omar S. M. El Nahhas\inst{2} \and
Tianyu Han\inst{1} \and
Gustav Müller-Franzes\inst{1} \and
Sven Nebelung\inst{1} \and
Jakob Nikolas Kather\inst{2, 3, 4} \and
Daniel Truhn\inst{1} 
}
%

%\author{Anonymous\inst{1} \and
%Anonymous\inst{2} \and
%Anonymous\inst{3} \and
%Anonymous\inst{1} \and
%Anonymous\inst{1} \and
%Anonymous\inst{4} \and
%Anonymous\inst{1}
%}
%
% TODO: ADD F.Khader et al. back in here
\authorrunning{F. Khader et al.}
%\authorrunning{Anonymous}

% First names are abbreviated in the running head.
% If there are more than two authors, 'et al.' is used.
%
% TODO: Comment this back in
\institute{
Department of Diagnostic and Interventional Radiology, University Hospital Aachen, Aachen, Germany \\ \email{fkhader@ukaachen.de}\and
Else Kroener Fresenius Center for Digital Health, Medical Faculty Carl Gustav Carus, Technical University Dresden, Dresden, Germany \and
Department of Medicine 1, University Hospital and Faculty of Medicine Carl Gustav Carus, TUD Dresden University of Technology, Germany \and
Medical Oncology, National Center for Tumor Diseases (NCT), University Hospital Heidelberg, Heidelberg, Germany 
}

%\institute{
%Anonymous Organization \and
%Anonymous Organization \and
%Anonymous Organization \and
%Anonymous Organization
%}

\maketitle              % typeset the header of the contribution
\begin{abstract}
The Transformer model has been pivotal in advancing fields such as natural language processing, speech recognition, and computer vision. However, a critical limitation of this model is its quadratic computational and memory complexity relative to the sequence length, which constrains its application to longer sequences. This is especially crucial in medical imaging where high-resolution images can reach gigapixel scale. Efforts to address this issue have predominantely focused on complex techniques, such as decomposing the softmax operation integral to the Transformer's architecture. This paper addresses this quadratic computational complexity of Transformer models and introduces a remarkably simple and effective method that circumvents this issue by eliminating the softmax function from the attention mechanism and adopting a sequence normalization technique for the key, query, and value tokens. Coupled with a reordering of matrix multiplications this approach reduces the memory- and compute complexity to a linear scale. We evaluate this approach across various medical imaging datasets comprising fundoscopic, dermascopic, radiologic and histologic imaging data. Our findings highlight that these models exhibit a comparable performance to traditional transformer models, while efficiently handling longer sequences. 
%The source code is made available for public access on GitHub at X.

\keywords{Attention  \and Transformers \and Medical Imaging}
\end{abstract}
\section{Introduction}
With the development of transformer-based models significant advancements in the field of machine learning have been witnessed in recent years \cite{vaswani_attention_2017}. Vision Transformers (ViTs) \cite{dosovitskiy_image_2021} have emerged as an effective method for processing imaging data, utilizing the attention mechanism to enhance object detection, segmentation and image classification in medical image analysis \cite{shamshad_transformers_2023}. Moreover, the capability to treat data from various modalities in an input-agnostic fashion enables an intuitive integration of multimodal data into the model's inference process \cite{khader_multimodal_2023}. However, a core limitation of self-attention based ViT models is their quadratic computational and memory demands as the input sequence length increases \cite{dosovitskiy_image_2021}, a problem that is exacerbated in the context of medical imaging where data often come high-resolution and multi-dimensional formats, spanning from 2D and 3D to even 4D (i.e., 3D + t) data. One example is histopathological imaging, where whole-slide images reach gigapixel scale, resulting in input sequence lengths that can exceed $10,000$ tokens, thus requiring expensive hardware to run \cite{khader_cascaded_2024}. 
\newline\newline
To address the scalability issue, multiple approaches aimed at reducing the quadratic scaling of the attention mechanism have been published in recent years  \cite{xiong_nystromformer_2021,choromanski_rethinking_2022,katharopoulos_transformers_2020,lu_soft_2021,qin_cosformer_2022,koohpayegani_sima_2024}. The primary focus of these methods has been to decompose the softmax operation and to reorder the subsequent matrix multiplications. For instance, Xiong et al. propose to approximate the attention mechanism using the Nyström technique \cite{xiong_nystromformer_2021}, involving a singular value decomposition to decompose the softmax. Choromanski et al. estimate the attention matrix through orthogonal random projections \cite{choromanski_rethinking_2022}. Moreover, Lu et al. propose to replace the softmax operation with a Gaussian kernel \cite{lu_soft_2021}, whereas Katharopoulos et al. suggest omitting the softmax by projecting the keys and queries using feature maps given by $\phi(x) = \text{elu}(x) + 1$, where $\text{elu}(.)$ denotes the exponential linear unit \cite{katharopoulos_transformers_2020}. Similarly, Qin et al. propose to use a feature map given by $\phi(x) = \text{ReLU}(x)$ and add a $cos$-based re-weighting mechanism to enforce stronger dependencies on neighboring tokens, as this bias is commonly observed in natural language processing (NLP) \cite{qin_cosformer_2022}. However, Koohpayegani et al. argue that the usage of kernel feature maps that involve the use of $\exp(.)$ operations can lead to inefficiencies on edge-devices as these are costly to implement \cite{koohpayegani_sima_2024}. This could hold significance in the field of medical imaging as data protection guidelines may require on-site solutions of AI models, thus emphasizing the need for affordable hardware solutions. Consequently, concurrent to our work, the aim of their study was to replace the softmax operation with a SimA block, which consists of a mere normalization of tokens, thus ensuring a simple implementation that does not suffer from inefficiencies caused by the $\exp(.)$ operation. Similarly to our work, they focus their application to vision recognition rather than NLP. However, while the authors employ an $\ell_1$ normalization across the key and query tokens, we propose the usage of a learnable normalization across the keys, queries and values. Moreover, their evaluation is limited to the ImageNet dataset for classification purposes. In contrast, our study addresses the broader area of medical image classification, demonstrating the versatility and applicability of our method across various types of medical imaging data.
\newline\newline
Therefore, our key contributions include: (1) The development of a simple softmax-free approach that involves a learnable normalization over the sequence dimension, achieving performances on par with traditional ViT models, while effectively handling data with longer sequences. (2) Showcasing the effectiveness of our technique across a diverse spectrum of medical imaging data, encompassing a variety of resolutions and sequence lengths. (3) Conducting a comparative analysis with the SimA method, highlighting our method's enhanced performance across a wide array of medical imaging datasets. 

\section{Materials \& Methods}
\subsection{Softmax-Free Transformer with Sequence Normalization}
\label{ssec:softmax_free}
Our method is based on the original attention mechanism as proposed by Vaswani et al. \cite{vaswani_attention_2017} and later adopted by Dosovitskiy et al. in the ViT model \cite{dosovitskiy_image_2021}: Given an image $\mathbf{x} \in \mathbb{R}^{D \times H \times W \times C}$, where $D$ denotes the image depth ($= 1$ for 2D images), $H$ denotes the image height, $W$ denotes the image with and $C$ the number of channels, it is first transformed into a sequence of patches $\mathbf{x}_p \in \mathbb{R}^{N \times (P_D \cdot P_H  \cdot P_W \cdot C)}$, where $P_D$, $P_H$, and $P_W$ denote the depth, height and width of the image patches and $N = \frac{D\cdot H\cdot W}{P_D \cdot P_H \cdot P_W}$ denotes the sequence length. Subsequently, each token in the sequence is projected onto a latent dimension $D'$ and a learnable \texttt{[class]} token is prepended to the input sequence (which we will overlook for simplicity). Moreover, a positional encoding is applied to each token in the input sequence to provide positional information about their location. To extract a class score, the resulting sequence undergoes processing through a series of multi-head self-attention layers, multilayer Perceptrons and Layernorm \cite{dosovitskiy_image_2021}. As our method ultimately focuses on modifying the self-attention layers in the ViT model, while keeping the overall architecture the same, our following discussion will solely focus on these layers. Let $\mathbf{x}_p^{(i)} \in \mathbb{R}^{N \times D'}$ denote the input to an arbitrary attention layer $i$, and let $W^Q \in \mathbb{R}^{D'\times D}$, $W^K \in \mathbb{R}^{D'\times D}$ and $W^V \in \mathbb{R}^{D' \times D}$ denote projection matrices that map the sequence of input tokens to matrices containing the queries $Q \in \mathbb{R}^{N \times D}$, keys $K \in \mathbb{R}^{N \times D}$ and values $V \in \mathbb{R}^{N \times D}$ in which each row contains the concatenated query, key and value representation of dimension $\frac{D}{J}$ of each head. Here, $J$ denotes the number of heads. The attention mechanism, as defined by Vaswani et al. \cite{vaswani_attention_2017}, is then expressed as
\begin{equation}
    Attention(Q_j,K_j,V_j) = \text{softmax}(\frac{Q_jK_j^T}{\sqrt{\frac{D}{J}}})V_j
\end{equation}
where $Q_j$, $K_j$ and $V_j$ denote the matrices containing the queries, keys and values belonging to a single head $j \in 1 ... J$. This computation exhibits a quadratic scaling of the attention mechanism with respect to the input sequence length, as the multiplication $Q_jK_j^T$ results in a matrix of dimensions $N \times N$. Note however, that if we omit the softmax computation one could exploit the associative property of matrix multiplications by first performing the computation $M_j=K_j^TV_j$, where $M_j$ is of dimensions $\frac{D}{J} \times \frac{D}{J}$ and subsequently performing the computation $Q_jM_j$, followed by a multiplication with the scaling term $\frac{1}{\sqrt{\frac{D}{J}}}$ \cite{choromanski_rethinking_2022}. In contrast to the previous approach, no matrix of dimension $N \times N$ has the be stored, nor computed. Therefore if $\frac{D}{J} < N$ such approach results in an improved scaling behavior. Building on this intuition, we employ a remarkably simple approach that involves omitting the softmax in the attention computation and instead normalizes the keys, queries and values such that the summation over attention-weighted value vectors does not increase to arbitrarily high values for extensive sequence lengths $N$. More precisely, we normalize each concatenated token representation $Q$, $K$ and $V$ for each batch element $b$ and feature element $f$ to have a mean value of zero and a variance of one
\begin{align}
    \hat{Q}^{(b, f)} &= \frac{Q^{(b, f)} - \mathrm{E}[Q^{(b, f)}]}{\sqrt{\mathrm{Var}[Q^{(b, f)}]}} \\
    \hat{K}^{(b, f)} &= \frac{K^{(b, f)} - \mathrm{E}[K^{(b, f)}]}{\sqrt{\mathrm{Var}[K^{(b, f)}]}} \\
    \hat{V}^{(b, f)} &= \frac{V^{(b, f)} - \mathrm{E}[V^{(b, f)}]}{\sqrt{\mathrm{Var}[V^{(b, f)}]}}
\end{align}
Moreover, we scale and shift the normalized values using learnable parameters $\gamma^{(f)}$ and $\beta^{(f)}$ separately for each key, value and query, following the batch normalization technique \cite{ioffe_batch_2015}, i.e.,:
\begin{align}
    \hat{Q'}^{(b, f)} &= \gamma^{(f)}\hat{Q}^{(b, f)} + \beta^{(f)} \\
    \hat{K'}^{(b, f)} &= \gamma^{(f)}\hat{K}^{(b, f)} + \beta^{(f)} \\
    \hat{V'}^{(b, f)} &= \gamma^{(f)}\hat{V}^{(b, f)} + \beta^{(f)}
\end{align}
This normalization resembles an instance normalization \cite{ulyanov_instance_2017}, where the normalization occurs for each feature channel separately across the sequence length. To highlight that this normalization occurs across the sequence dimension, we refer to this normalization as sequence normalization in the following. Additionally, we replace the scaling term of the original approach by $\frac{1}{N}$ to reduce the dependence of the subsequent vector sum on the input sequence length.

\begin{figure}[h]
\centering
\includegraphics[width=0.85\textwidth]{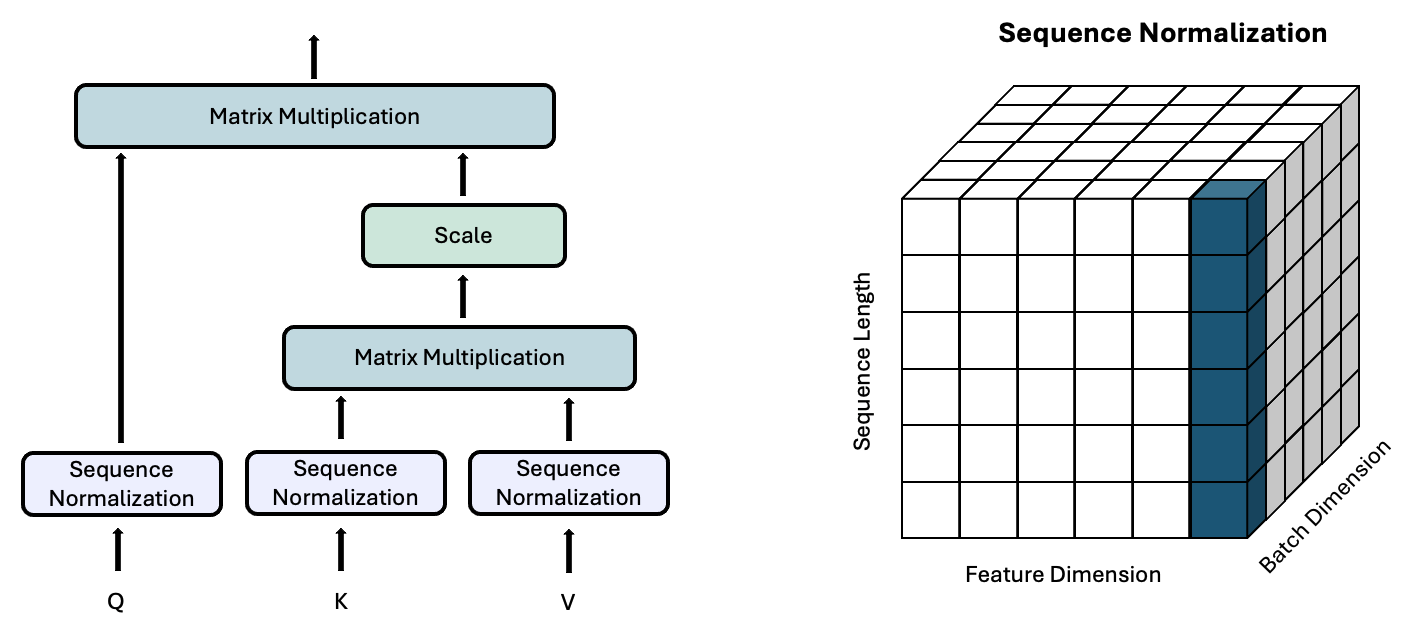}
\caption{Illustration of the attention computation mechanism (left) and sequence normalization approach (right) used in our study. Note that sequence normalization resembles an instance normalization step across the sequence dimension.} 
\end{figure}

\subsection{Datasets}
We demonstrate the effectiveness of our method on a wide range of publicly available medical imaging datasets, including both 2D and 3D images. Detailed descriptions of these datasets are provided below.

\subsubsection{Fundoscopic Images}
We train the models on a dataset containing fundoscopic imaging data extracted from the AIROGS challenge \cite{de_vente_airogs_2024}, which consists of 101,442 images with labels denoting the presence of referable glaucoma. We split this dataset into a training set comprising n=64,922 images, a validation set with n=16,231 images and a test set comprising 20,289 images. All images were resized to a resolution of $224\times224$ pixels.

\subsubsection{Dermascopic Images}
Moreover, the models were trained on the ISIC dataset \cite{rotemberg_patient-centric_2021}, which contains 33,126 dermascopic images of benign and malignant skin lesions from 2,056 patients. As before, we split this dataset into a training (n=21,138), validation (n=5,273) and test (n=6,715) set. As before, all images were resized to a resolution of $224\times224$ pixels.

\subsubsection{Chest Radiographs}
Additionally, we trained the models on a dataset comprising 15,000 chest radiographs, which were extracted from the official training set of the VinDR-CXR dataset \cite{nguyen_vindr-cxr_2022}. Since each image was rated by multiple radiologists, the consensus rating served as the ground truth. All models were trained to predict the presence or absence of cardiomegaly. The dataset was split into a training (n=9,000), validation (n=3,000) and test set (n=3,000). To demonstrate the performance of our model on different input resolutions, we trained the model on an image resolution of $224\times224$ pixels, resulting in 196 patches when a patch size of (16, 16) is employed, and a resolution of $1024\times1024$ pixels, resulting in 4096 patches.

\subsubsection{Breast MRI Scans}
We utilized a publicly available dataset of 3D breast cancer MRI scans acquired from a population of 922 female patients \cite{saha_dynamic_2021}. We trained the models on subtraction images that were created by subtracting the T1-weighted fat-suppressed post-contrast scans from the pre-contrast scans. Each image was independently resampled to a voxel size of $0.7\times0.7\times3 \text{mm}$ and bilaterally split along the median plane. Moreover, the images were center-cropped to a resolution of $256\times256\times32$ voxels. To classify the cancer status, patients for which the laterality of breast cancer was unknown (n=271) were not considered in our analysis. This results in a dataset of 1302 samples that were further split into a training (n=834), validation (n=208) and test (n=260) set. We will refer to this dataset as DUKE in the following.  

\subsubsection{Whole-Slide Images}
To demonstrate the performance on tasks in which the input sequence length can become extensive, we trained the models on histological whole-slide images (WSIs) comprised in the TCGA-NSCLC (non-small-cell lung cancer) dataset \cite{khader_cascaded_2024}. This dataset is comprised of two sub-datasets: TCGA-LUSC (lung squamous cell carcinoma, n=512) and TCGA-LUAD (lung adenocarcinoma, n=530). The dataset was split into a training set (n=627), a validation set (n=145) and a test set (n=270) and the models were trained on the task of cancer subtyping. The images were preprocessed by extracting square patches of \SI{256}{\micro\metre}, resizing them to a resolution of $256\times256$ pixels and dropping uninformative patches (i.e., those that are pre-dominantly white or blurry). Moreover, patch-based feature representations were extracted using RetCCL \cite{wang_retccl_2023}, resulting in sequence lengths of up to 11,039 tokens.

\section{Experiments \& Results}
\subsection{Neural Network Architecture}
We benchmarked our approach using the ViT model across diverse medical imaging datasets (as detailed in the previous section). Table \ref{tab:hyperparameters} details the hyperparameters for both 2D and 3D models. We train the 2D models on images of size $224 \times 224$ with a batch size of 32 and images of size $1024 \times 1024$ with a batch size of 4. For WSIs a batch size of 1 is employed to accommodate the available GPU VRAM, while models trained with 3D imaging data utilize a batch size of 16. All models were trained on an NVIDIA RTX A6000 GPU for 100 epochs. For each dataset, we train and evaluate three distinct models: (1) A vanilla ViT model; (2) our proposed model replacing the traditional ViT attention mechanism with the softmax-free approach outlined in Section \ref{ssec:softmax_free}; and (3) the SimA approach \cite{koohpayegani_sima_2024} for comparison. Data augmentation for 2D imaging datasets (excluding whole-slide images) involves rotations by $\pm 45 \degree$ and vertical flipping with a 50\% chance. For the 3D imaging data we apply flipping across all spatial dimensions with a 50\% probability. 

\begin{table}[htbp]
\centering
\caption{Hyperparameters used to train the ViT models in our study. ViT\textsubscript{2D} was used for 2D images, ViT\textsubscript{WSI} was used for whole-slide images and ViT\textsubscript{3D} for 3D images.}
\label{tab:hyperparameters}
\begin{tabular}{llll}%{l@{\hspace{0.8cm}}l@{\hspace{0.8cm}}l@{\hspace{0.8cm}}l}
\toprule
\textbf{Parameter}      & \textbf{ViT\textsubscript{2D}} & \textbf{ViT\textsubscript{WSI}} & \textbf{ViT\textsubscript{3D}} \\ 
\midrule
\textbf{No. Channels}   & 3            & -                                 & 1              \\
\textbf{Patch Size}     & (16, 16)     & -                                 & (16, 16, 4)    \\
\textbf{$D'$}           & 1024         & 512                               & 1024           \\
\textbf{$D$}            & 512          & 512                               & 512            \\
\textbf{No. Layers}     & 8            & 2                                 & 8              \\
\textbf{No. Heads}      & 8            & 8                                 & 8              \\
\textbf{MLP Dim}        & 1024         & 512                               & 1024           \\
\bottomrule
\end{tabular}
\end{table}

\subsection{Image Classification}
We evaluate the effectiveness of the softmax-free transformer model across a diverse array of image classification tasks. For each model and dataset we assess the performance based on the area under the receiver operating characteristic (AUROC) score. Remarkably, our findings reveal that the performance of our proposed softmax-free model is on par with that of the traditional ViT model across all datasets, with the exception of the fundoscopic data. Here, our approach performance slightly worse than the ViT model. Furthermore, our analysis demonstrates that, in comparison to the SimA model, our softmax-free approach exhibits superior performance across all five datasets employed for evaluation. Table \ref{tab:modelPerformance} details the comparative performance metrics for all five datasets. 
\begin{table}[htbp]
\centering
\caption{Comparison of AUROC scores across all datasets and models.}
\label{tab:modelPerformance}
\begin{tabular}{lccc}%{
    %l 
    %@{\hspace{0.8cm}} S[table-format=1.2] @{\hspace{0.8cm}} 
    %@{\hspace{0.8cm}} S[table-format=1.2] @{\hspace{0.8cm}} 
    %@{\hspace{0.8cm}} S[table-format=1.2] 
%}
\toprule
\textbf{Dataset} & \textbf{ViT} & \textbf{SimA} & \textbf{Ours} \\
\midrule
DUKE & 0.77 & 0.71 & $\mathbf{0.84}$ \\
VinDr-CXR ($224\times224$) & 0.86 & 0.86 & $\mathbf{0.87}$ \\
VinDr-CXR ($1024\times1024$) & $\mathbf{0.89}$ & $0.77$ & $\mathbf{0.89}$ \\
AIROGS & $\mathbf{0.89}$ & 0.73 & $0.83$ \\
ISIC & 0.86 & 0.87 & $\mathbf{0.88}$ \\
TCGA-NSCLC & $\mathbf{0.97}$ & 0.94 & $\mathbf{0.97}$ \\
\bottomrule
\end{tabular}
\end{table}

\subsection{Scaling Behavior}
\begin{figure}[htb]
\label{fig: scaling}
\centering
\includegraphics[width=0.88\textwidth]{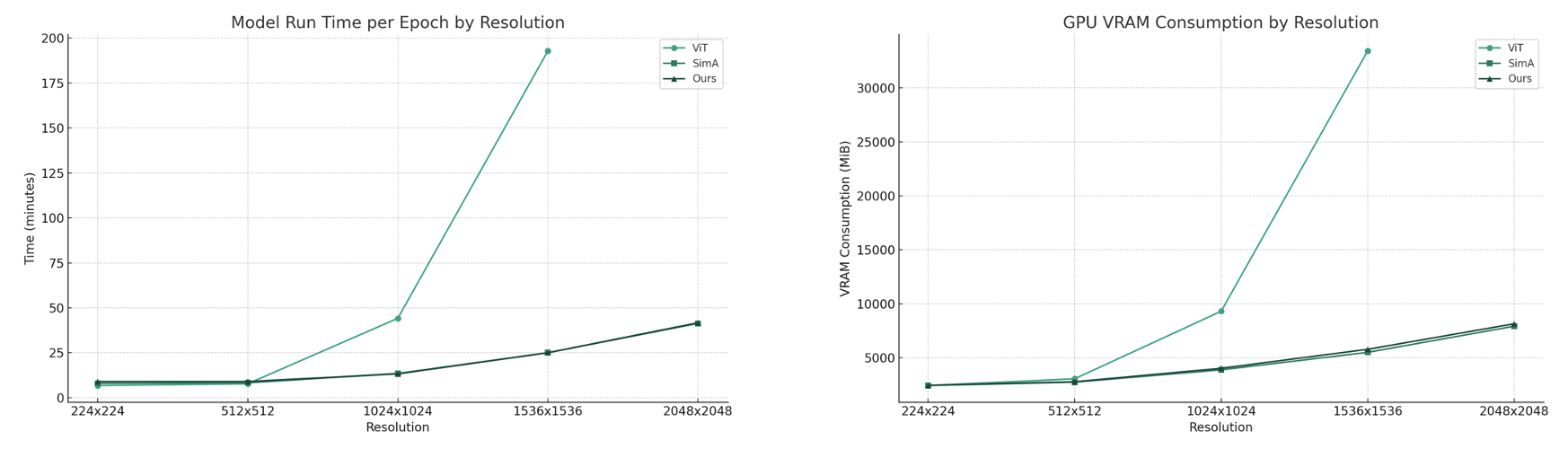}
\caption{Scaling behavior of the three models with respect to the input resolution. We find that our approach, as well as the SimA model scale much more efficiently in terms of epoch duration time (left) as well as GPU VRAM consumption (right) for the 15,000 images contained in the VinDr-CXR dataset (left). Note that a batch size of 1 was utilized for all runs. Evaluations for the ViT at image resolutions of $2048\times2048$ were omitted as the model did not fit into the 48GB GPU memory.}
\end{figure}

To assess variations in scaling behavior with regard to compute- and memory usage, we conducted a series of training runs using all 15,000 images of the VinDr-CXR dataset and recorded the run time per epoch at multiple resolutions typically encountered in medical imaging. Moreover, we record the GPU VRAM consumption for training with a batch size of $1$. These resolutions range from $224\times224$ to $2048\times2048$. Given a patch size of $16\times16$ these resolutions correspond to sequence lengths between 196 and 16,384. The latter also serves as a good approximation for what is typically required when training on histological whole-slide images, which often encompass more than 10,000 input tokens. We found that for both, our approach and SimA, the scaling behavior is much more beneficial compared to the traditional ViT: While training time and memory consumption quickly escalated for the ViT, the softmax-free approaches showed benevolent behavior even for input images of size $2048\times2048$. For additional insights into the scaling behavior, please refer to Figure \ref{fig: scaling}.

\subsection{Softmax-Free Finetuning}
To improve the performance on the AIROGS dataset, we conducted an ablation study to evaluate the impact of fine-tuning a pre-trained vanilla ViT model with our softmax-free approach. Specifically, we initialized softmax-free model with the model weights corresponding to the checkpoint that led to the highest validation score for the vanilla ViT model. We then trained this model for 100 epochs. Our finding reveal that this approach improves the AUROC score for the softmax-free model to 0.84, suggesting that our softmax-free approach is effective in adopting feature representations learned by conventional transformer models.

\section{Conclusion}
In this paper we have presented a remarkably simple approach for achieving linear scaling behavior with transformer models. We demonstrated the capabilities of this approach on a diverse set of publicly accessible medical imaging data that covers a wide array of different modalities. Our findings reveal that our softmax-free approach enables us to achieve comparable performances to that of traditional softmax-based ViT models, while allowing for faster computations and reduced memory consumption when sequence lengths become extensive. Furthermore, our approach outperforms the recently proposed SimA model, particularly in scenarios involving higher resolutions and more complex modalities characterized by longer sequence lengths, while being equally simple to implement and not requiring $exp(.)$ operations, a crucial aspect for fast edge-device computations. This suggests that attention-based models can be efficiently used in environments with limited computing resources, potentially broadening their application in various clinical settings. Although we have empirically validated the efficacy of our approach, future investigations should delver deeper into the workings of such softmax-free methods and further validate them across additional datasets to understand their limitations.

\subsubsection{Acknowledgements.}
The results published here are in whole or part based upon data generated by the TCGA Research Network: \url{https://www.cancer.gov/tcga}. This research is funded by the German Federal Ministry of Education (TRANSFORM LIVER, 031L0312A).

%
% ---- Bibliography ----
%
% BibTeX users should specify bibliography style 'splncs04'.
% References will then be sorted and formatted in the correct style.
%
% \bibliographystyle{splncs04}
% \bibliography{mybibliography}
%
\bibliographystyle{splncs04}
\bibliography{references2}

\end{document}